\documentclass[journal,twoside,web]{ieeecolor}
\newif\ifTrackChanges   
\TrackChangestrue 
\definecolor{deep-red}{RGB}{192, 0, 0}
\definecolor{deep-purple}{RGB}{120, 0, 170}
\definecolor{good-green}{RGB}{0,175,0} 
\definecolor{purple}{RGB}{210, 0, 210} 
\definecolor{alizarin}{rgb}{0.82, 0.1, 0.26}

\usepackage[textsize=tiny, bordercolor=white,backgroundcolor=gray!30,linecolor=black,colorinlistoftodos]{todonotes}  
\usepackage[normalem]{ulem} 
\usepackage{soul} 

\usepackage[mathscr]{euscript}
\ifTrackChanges

    \newcommand{\cut}[1]{{\color{gray}{#1}}}
\else

    \newcommand{\cut}[1]{{}}

\fi

\usepackage{generic}
\usepackage{cite}
\usepackage{amsmath,amssymb,amsfonts}
\usepackage{algorithm}
\usepackage{algorithmic}
\usepackage{graphicx}
\usepackage{textcomp}
\usepackage{dblfloatfix}
\usepackage{marginnote}
\usepackage{soul}
\usepackage{hyperref}
\usepackage{tabularx}
\usepackage{multirow}
\usepackage{boldline}
\usepackage{nicematrix}
\usepackage{fancyhdr}
\hypersetup{
  colorlinks   = true, 
  urlcolor     = blue, 
  linkcolor    = blue, 
  citecolor   = green 
}
\newcommand*\circled[1]{\tikz[baseline=(char.base)]{\node[circle,minimum size=7pt,draw=black,inner sep=0.5pt](char){\scriptsize #1};}}
\def\BibTeX{{\rm B\kern-.05em{\sc i\kern-.025em b}\kern-.08em
    T\kern-.1667em\lower.7ex\hbox{E}\kern-.125emX}}
\begin{document}
\title{Endovascular Detection of Catheter-Thrombus Contact by Vacuum Excitation}
\author{Jared Lawson$^{1}$, Madison Veliky$^{1}$, Colette P. Abah$^{2}$, Mary S. Dietrich$^{3}$, Rohan Chitale$^{4}$, and Nabil Simaan$^{1,\dag}$
\thanks{J. Lawson was supported by NIH award \#T32EB021937 and by Vanderbilt University funds.}
\thanks{$\dag$ Corresponding author}
\thanks{$^{1}$Department of Mechanical Engineering, Vanderbilt University, Nashville, TN 37235, USA
        {\tt\small (jared.p.lawson, madison.a.veliky, nabil.simaan) @vanderbilt.edu}}%
\thanks{$^{2}$Intuitive Surgical, Sunnyvale, CA 94086, USA
        {\tt\small (colette.abah) @gmail.com}}%
\thanks{$^{3}$Department of Biostatistics, Vanderbilt University Medical Center, Nashville, TN 37235, USA
        {\tt\small (mary.dietrich) @vumc.org}}%
\thanks{$^{4}$Department of Neurological Surgery, Vanderbilt University Medical Center, Nashville, TN 37235, USA
        {\tt\small (rohan.chitale) @vumc.org}}}
\maketitle

\thispagestyle{fancy}
\fancyhf{}
\renewcommand{\headrulewidth}{0pt}
\lhead{2024 IEEE Transactions on Biomedical Engineering. Accepted Version. }
\rfoot{\centering \scriptsize \copyright 2024 IEEE. Personal use of this material is permitted. Permission from IEEE must be obtained for all other uses, in any current or future media, including reprinting/republishing this material for advertising or promotional purposes, creating new collective works, for resale or redistribution to servers or lists, or reuse of any copyrighted component of this work in other works.}
\begin{abstract}
Objective: The objective of this work is to introduce and demonstrate the effectiveness of a novel sensing modality for contact detection between an off-the-shelf aspiration catheter and a thrombus.  \\
Methods: A custom robotic actuator with a pressure sensor was used to generate an oscillatory vacuum excitation and sense the pressure inside the extracorporeal portion of the catheter. Vacuum pressure profiles and robotic motion data were used to train a support vector machine (SVM) classification model to detect contact between the aspiration catheter tip and a mock thrombus. Validation consisted of benchtop accuracy verification, as well as user study comparison to the current standard of angiographic presentation.\\
Results: Benchtop accuracy of the sensing modality was shown to be 99.67\%. The user study demonstrated statistically significant improvement in identifying catheter-thrombus contact compared to the current standard. The odds ratio of successful detection of clot contact was 2.86 (p=0.03) when using the proposed sensory method compared to without it. \\
Conclusion: The results of this work indicate that the proposed sensing modality can offer intraoperative feedback
to interventionalists that can improve their ability to detect contact between the distal tip of a catheter and a thrombus.  \\
Significance: By offering a relatively low-cost technology that affords off-the-shelf aspiration catheters as clot-detecting sensors, interventionalists can improve the first-pass effect of the mechanical thrombectomy procedure while reducing procedural times and mental burden.
\end{abstract}

\section{Introduction} \label{sec:introduction}
\IEEEPARstart{I}{schemic} stroke has an immense impact on the global population, with over 7 million incidents reported in 2020 \cite{tsao2023heart}. Approximately 30\% of acute ischemic strokes are caused by large vessel occlusions (LVOs), in which a clot located in a proximal cerebral vessel prohibits blood flow to a significant portion of brain tissue \cite{Lakomkin2019}. LVOs lead to devastating patient outcomes \cite{smith2006prognostic,smith2009significance} and account for over 90\% of post-stroke mortality \cite{malhotra2017}. Every second of ischemia is detrimental to patient outcomes, and emergent treatment is necessary.
\par Mechanical thrombectomy (MT) has become a standard-of-care for emergent LVO treatment, in which endovascular devices are navigated to the thrombus, which is retrieved by aspiration and/or stent-retriever, ideally recanalizing the vessel. While multiple clinical trials have shown that MT offers patients the best clinical outcomes compared to other treatments \cite{Berkhemer2015,Campbell2015,Goyal2015,Jovin2015,Saver2015}, the procedure remains challenging and must be further improved. 
\par Positive patient outcomes following MT are directly correlated with how quickly the interventionalist is able to recanalize the occluded vessel \cite{zaidat2018first}. A related factor is first-pass effect (FPE), which refers to achieving successful clot retrieval on the first attempt. FPE serves as a measure of procedural success since it correlates with best patient outcomes, although it is achieved in only one-third of thrombectomy cases \cite{zaidat2018first,abbasi2021FPE}. To achieve FPE during aspiration thrombectomy (otherwise known as ADAPT \cite{ADAPT2014}), a \textit{necessary condition} is proper engagement between the thrombus and the distal tip of the catheter \textit{prior} to attempting aspiration. Aspiration without such engagement results in failed retrieval attempts and increases both ischemic time and procedural cost. Details about the challenges of detecting this contact are further elaborated in Section \ref{sec:motivation}. This paper aims to present a low-cost sensory technology that can ensure proper detection of this catheter-thrombus engagement. 
\begin{figure}[htbp]
        \centering
        \includegraphics[width=0.99\columnwidth]{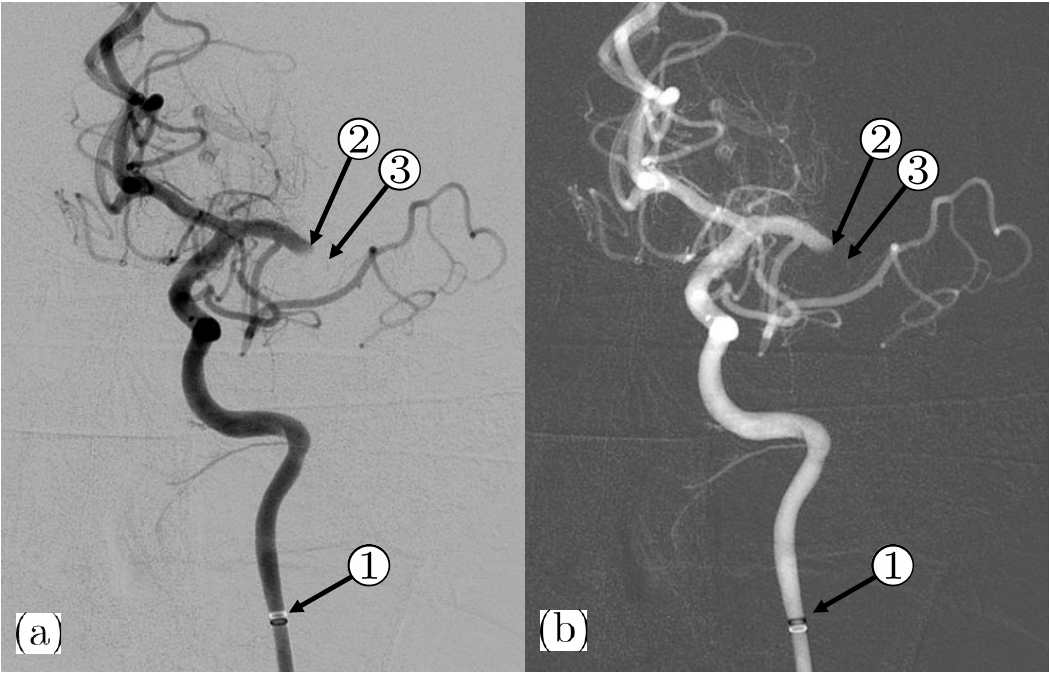}
        \caption{M1 Segment MCA Occlusion Angiogram. (a) is the true digital subtraction angiogram (DSA), while (b) is the roadmap of the DSA used for navigation guidance. \protect\circled{1} radiopaque catheter tip as viewed by the interventionalist, \protect\circled{2} location of halting contrast flow, \protect\circled{3} expert estimated location of the true thrombus location}\label{fig:real_occlusion}
\end{figure}
\par The contribution of this paper is in presenting a novel method for determining contact between an aspiration catheter with a thrombus. This method, of which a preliminary version was presented in \cite{Veliky2023}, relies on using robotically-forced vacuum excitation and sensory monitoring of pressure at the extracorporeal (proximal) end of the aspiration catheter. Relative to existing technologies for endovascular sensing, the proposed method does not require specialized sensing instrumentation or modification of the aspiration catheter in any way. Therefore, our proposed method is suitable for very small catheters with minimal disruption to the clinical workflow. 
\par The significance of our proposed sensory modality is shown in an ex-vivo user study where expert neurointerventionalists' ability to detect engagement between their catheter tip and a thrombus are substantially improved using this approach compared with the current standard-of-care; thereby minimizing the chance of a failed aspiration attempt and increasing the chances of achieving FPE. 
\par In the next section, we present the clinical motivation driving this research. Section \ref{sec:relevantworks} reviews relevant works related to sensing modalities and clarifies current gaps in the field. In Section \ref{sec:methods}, the current and proposed clinical workflows are introduced to motivate the sensing technology, which is then described in detail. Specifically in Sections \ref{sec:expmethod} and \ref{sec:studymethod}, the protocols for benchtop experimental validation and the user study are presented. Section \ref{sec:results} includes results collected with the system to verify the accuracy of the approach (Sec.~\ref{sec:experiments}), followed by results of the user study with expert neurointerventionalists (Sec.~\ref{sec:userstudy}). Study outcomes and future works are discussed in Section \ref{sec:discussion}, and key takeaways are summarized in Section \ref{sec:conclusion}.
%
\subsection{Clinical Motivation}\label{sec:motivation}
Fig.~\ref{fig:real_occlusion}(a) shows a digital subtraction angiogram (DSA) of vasculature with an occlusion in the M1 segment of the middle cerebral artery (MCA). The negative of the DSA, shown in Fig.~\ref{fig:real_occlusion}(b), is then used as a fixed navigation roadmap which shows the vasculature and provides a visual \textit{guess} of the location of the thrombus \cite{Turski1982,consoli2018FPE,monch2021angiographic}. This roadmap of the DSA is superimposed over subsequent fluoroscopy images, which clearly show the radiopaque catheter tip (indicated by \circled{1} in Fig.~\ref{fig:real_occlusion})\cite{Turski1982}.
\par While this clinical technique is helpful in providing a visual aid for catheter or guidewire navigation, it does not exactly indicate the thrombus location. In current practice, interventionalists estimate the thrombus location by observing the halt of contrast agent into a vessel, which is indicated by \circled{2} in Fig.~\ref{fig:real_occlusion}(a)-\ref{fig:real_occlusion}(b). The halt of contrast does not always accurately indicate the thrombus location, as standing columns of contrast may not allow flow directly to the thrombus surface. An occluded vessel consists of a blood-filled, unoccluded region followed by an occluding thrombus. When contrast agent is injected into the vessel towards the thrombus, the incompressibility of the blood in the unoccluded region may prohibit the contrast agent from reaching the thrombus surface. Under angiography, this presents as a more proximally-located thrombus even if the true thrombus location is much deeper. Interventionalists are trained to wait for the contrast agent to diffuse with the hope of enough contrast agent concentration spreading to reach the thrombus, however they must also consider the contrast/visibility of the preceding vessel branches to ensure safe navigation.
\par Furthermore, the static DSA image used as a roadmap for navigation highlights the vessel before devices are inserted. As catheters and guidewires interact with the vessel wall, the true vessel location will deform from the roadmap image and thus further exacerbate the uncertainty of the thrombus location.
\par Because of the uncertainty regarding clot location, interventionalists must resort to approximating the aspiration catheter tip to the thrombus and judiciously deciding to apply suction when they think that the catheter tip has engaged the clot. If they guessed correctly, when they apply suction the catheter will not show blood return. However, absent blood flow as an indication for contact does not account for the ability of a blood vessel to collapse under aspiration pressure, which has been clinically reported in \cite{liu2022arterial}. Other methods using existing endovascular tools under fluoroscopy imaging alone have been proposed in case reports, both of which require penetrating the full clot with a microcatheter and flushing contrast distally \cite{nemoto2021pull,ohshima2020novel}. The penetration of a microcatheter into the thrombus risks further clot separation and distal embolization. Additional sensory feedback can potentially relieve this clinical challenge, and existing sensing modalities are reviewed in the following section.
\begin{figure*}[!b]
        \centering
        \vspace{0.5cm}
        \includegraphics[clip,trim=0cm 0.5cm 4cm 0cm,width=1.005\textwidth]{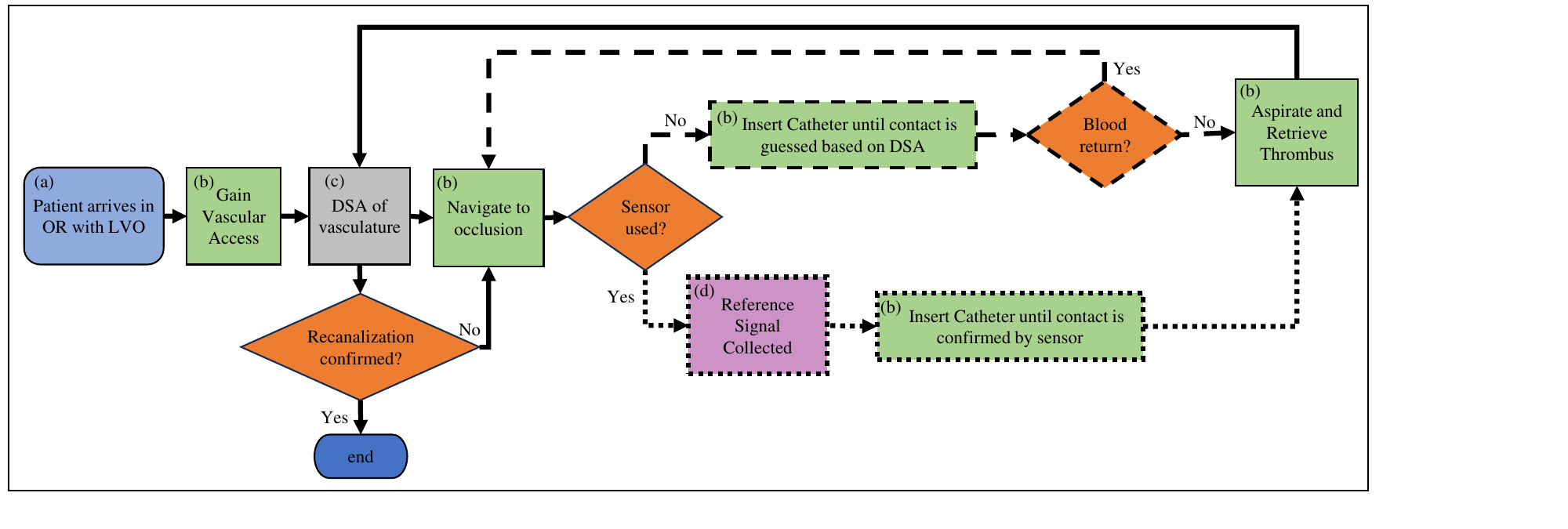}
        \caption{Clinical Workflow for Mechanical Thrombectomy with Aspiration (ADAPT \cite{ADAPT2014}). The first block (a) represents the starting point of the procedure. Blocks labeled (b) require manual navigation steps of the procedure. (c) blocks indicate where intraoperative imaging is used to visualize the anatomy. The (d) block and dotted lines indicate steps that are introduced with the proposed clinical workflow, which include sensing by vacuum excitation. Dashed lines refer to steps which are in the current workflow, but would not be  used during the proposed workflow}\label{fig:workflow}
\end{figure*}
%
\subsection{Relevant Works} \label{sec:relevantworks}
\par Preoperative imaging modalities, including computed tomography angiography (CTA) and magnetic resonance angiography (MRA), have been used for localization of thrombus before intervention is performed. Recently the use of machine learning has offered an opportunity for automated detection of LVOs from preoperative imaging \cite{amukotuwa2019automated,stib2020detecting,karamchandani2023automated}, however, these modalities fail to offer thrombus localization during the procedure.
\par For interventional procedures outside of the brain, catheters have been designed with sensing modalities built into the tip. Forward-viewing ultrasound transducers have been proposed for electrophysiological procedures for ablation with reduced fluoroscopic imaging \cite{Stephens2009}, and a 0.89mm (0.035") diameter robotic guidewire has more recently been proposed for peripheral vascular navigation of chronic total occlusions \cite{collins2020robotically}. Intravascular ultrasound (IVUS) probes have been designed within catheter tips for use in percutaneous coronary interventions, and have been used to characterize thrombus formation before placing coronary stents \cite{porto2004IVUS,vijayvergiya2021IVUS}. IVUS has been reportedly used in a similar fashion for intracranial and extracranial carotid stent placement \cite{hussain2016IVUS}. Doppler ultrasound transducers have been used to detect blood flow for coronary applications \cite{denardo1997doppler} and even cochlear blood flow \cite{ritter2000doppler}, and thus may show promise in detecting regions of abrupt changes in blood flow. Recently, fiberoptic angioscopes have been proposed for use in direct visualization of the neurovasculature \cite{Lazaro2019}. Ex-vivo experiments have demonstrated use of a 1.66mm (0.065", 5Fr) angioscope in visualizing plaque within carotid arteries \cite{savastano2017scope}, while another modified a ureteroscope for ex-vivo experiments in a carotid artery model \cite{zhang2022scope}. A combined tip force sensor and imaging probe were demonstrated for optical biopsy in the airway \cite{wu2020fbg}. Fagogenis et al. demonstrated distal image-guided navigation of cardiac catheters by classifying between instances of contact between the catheter and a vessel wall to aid in navigation \cite{fagogenis2019}. This work is particularly relevant, as the group employed SVM classification to detect such contact instances, however they classify images from distal catheter-tip imaging. Additional works related to tactile and perceptive sensing technologies for other clinical applications are reviewed in \cite{huang2020tactile}.
\par While these proposed catheter-tip sensing approaches are focused on different clinical applications other than aspiration thrombectomy, they each represent a modality which provides distal sensing under endovascular navigation. One of the major shortcomings of each tip-sensing modality in the context of aspiration thrombectomy is the inability to pass internal devices or aspirate through the sensing catheter. In order to effectively navigate deep into the vasculature, guidewires are often deployed through the catheter lumen to provide distal support. By prohibiting the passage of a guidewire by placing a distal tip sensor on the catheter, the interventionalists' task of navigation becomes more challenging if not impossible. Additionally, a distal tip sensor that eliminates a channel for aspiration or passage of stent-retrieval devices further challenges the clinical workflow and may increase the procedural time of MT. A sensory modality that eliminates the need for specialized sensing instrumentation embedded in the distal catheter tip represents an ideal approach for intraoperative catheter-thrombus contact detection. 
\section{Materials and Methods}\label{sec:methods}
\subsection{System Overview} \label{sec:system}
\subsubsection{Clinical Workflow of Aspiration Thrombectomy}
\par The current workflow for aspiration MT consists of elements of preoperative imaging, intraoperative imaging, and interventional device navigation and deployment. Figure~\ref{fig:workflow} shows the standard workflow (shown in dashed lines) and the sensory-guided workflow (shown in dotted lines).
\par Preoperative images are typically provided in the form of CTA or MRA. As MT is only recommended for patients with LVO, usually this preoperative imaging is required to confirm the region of the thrombus and the suitability of the patient for MT intervention. 
\par Once the patient is positioned in the interventional suite, vascular access is gained most commonly through the femoral artery, or less commonly through the radial artery. Navigation is carried out under the visualization of bi-plane fluoroscopy. A combination of guidewires, guide catheters, and intermediate catheters are used to navigate the major arteries of the head and neck, proximal to the predicted site of occlusion. This guide catheter is most commonly navigated to the internal carotid arteries (ICA), as they feed the most common site of LVOs in the middle cerebral arteries (MCA). At this point, a still DSA of the vessels of interest is taken by flushing a radiopaque contrast agent through this guide catheter. From this DSA, the interventionalist determines the estimated thrombus location by intuiting where the contrast should have, but did not flow. With this estimated thrombus location targeted, the interventionalist continues to navigate to the suspected occlusion site, taking additional DSA's as needed. Once near the occlusion site, the guidewire and intermediate catheters are removed from the guide catheter, and an aspiration catheter is inserted and approximated to the guessed thrombus location. Once contact of the distal catheter tip with the clot is guessed, aspiration is applied for 3-5 minutes to assist in seating the catheter tip into the clot. Subsequently, the catheter is removed. A post-thrombectomy DSA is then taken to confirm vessel recanalization. If the occlusion remains, the procedure needs to be repeated until successful or until further attempts are no longer meaningful. It is reported that favorable outcomes do not improve after three retrieval attempts due to elapsed ischemic time \cite{flottmann2018,alderson2022}.
\subsubsection{The Proposed Sensory-Guided Workflow}
\par The proposed catheter-thrombus contact detection system creates small vacuum (negative pressure) oscillations, or excitations, within the aspiration catheter, and monitors the resulting internal pressure signal. As a catheter creates vacuum excitations in an unoccluded blood vessel, blood freely flows and a measured internal pressure signal corresponds mostly with the flow head-loss. Once the catheter contacts the thrombus, vacuum excitations no longer correspond with back flow of blood into the catheter, therefore a markedly different internal pressure signal is observed. The system interprets such a marked difference in pressure signal by means of a support vector machine (SVM) classification algorithm that confirms or refutes a contact hypothesis with the clot. This classification algorithm is detailed in Section \ref{sec:algorithm}.
\par As seen in Fig.~\ref{fig:workflow}, the clinical workflow steps related to preoperative imaging, intraoperative imaging, or catheter navigation remain unaltered. However, as evident from the figure, the sensory-guided path can avoid the need for several aspiration attempts resulting in blood return due to a false guess of contact with the clot. 
\par According to the sensory-guided workflow, once the first DSA image is taken proximal to the occlusion site, at which point the aspiration catheter is inserted to a fully canalized, unoccluded portion of vessel, a 3 second reference vacuum excitation is performed to measure a baseline pressure wave. After this reference signal is recorded, the interventionalist continues inserting the aspiration catheter until the catheter tip is close to the clot. The sensory detection algorithm performs vacuum excitation for two seconds and provides an auditory signal informing the interventionalist whether or not the catheter tip is in contact with the thrombus. The interleaved sense/catheter advance steps are repeated as many times as necessary until catheter-thrombus contact is confirmed, at which point full aspiration may be applied as mentioned in the current clinical workflow.
\subsection{Vacuum Sensing System Overview}
\begin{figure}[htbp]
        \centering
        \includegraphics[width=0.98\columnwidth]{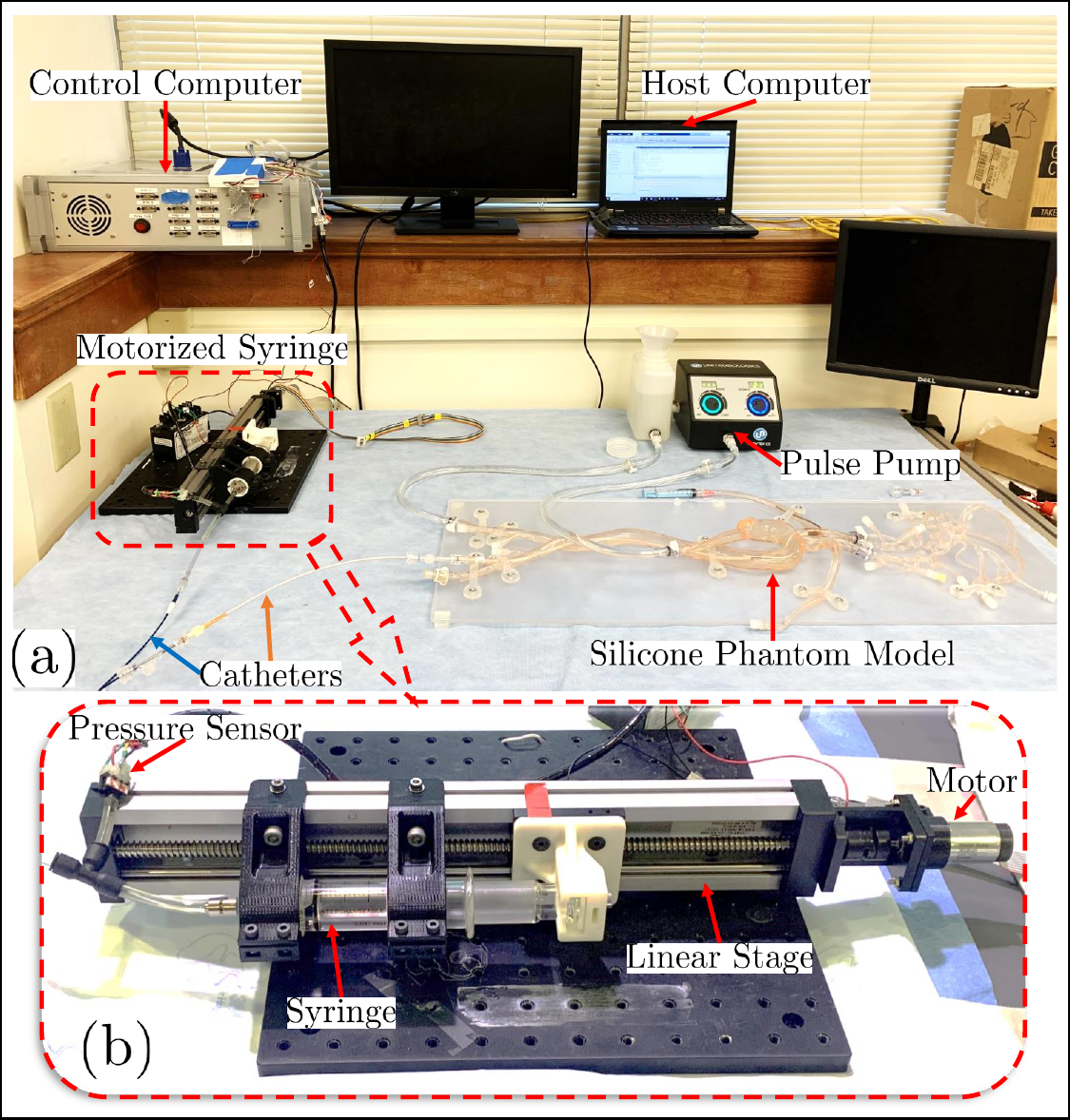}
        \caption{Sensing System Overview. (a) shows a higher-level image including the control computers, motorized syringe connecting to the aspiration catheter (left 'Catheters' arrow which is inserted in the guide catheter (top 'Catheters' arrow), which is inserted into the phantom model. (b) focuses on the motorized syringe to show the pressure sensor connected to the syringe output.}\label{fig:sensing_system}
\end{figure}
\par The system (see Fig.~\ref{fig:sensing_system}(a)) includes a 30mL syringe (FORTUNA, 7.140-44) as would be used in the existing clinical workflow, which is attached to a linear stage. The syringe body is fixed and the plunger motion is controlled to oscillate a stroke of 0.4mL at 4Hz. This stroke volume is chosen to minimize the ingested blood volume, and corresponds to around $1/8^{th}$ of the overall volume of the catheter. The frequency of the oscillation should ideally be higher than a standard heartrate, which would be maximum around 200bpm (3.33Hz). From tuning the motorized at 4Hz, 6Hz, and 8Hz, it was determined that 4Hz oscillations returned the most significant signal, which is likely due to the dynamic response of our design at the higher frequencies. The motorized stage (highlighted in Fig.~\ref{fig:sensing_system}(b)) includes a linear ball screw (McMaster Carr, 6734K34) and a DC motor with an encoder (Maxon, 369848). Between the syringe and the aspiration catheter (Medtronic, React 71), a pressure transducer (Honeywell, 26PCCFB2G) is connected to measure the pressure of the blood or saline flowing through the syringe-catheter assembly. The pressure transducer signals are amplified through a signal conditioner (Omega Engineering, DMD-465WD). The real-time control computer (Sensoray, Model 526) runs Simulink Real-Time at a 1kHz control frequency, commanding the motor position and reading the amplified pressure signals. A host computer is used to run the clot detection algorithm presented in Section \ref{sec:algorithm}, which communicates directly with the real-time control computer.  
\begin{figure}[htbp]
        \centering
        \includegraphics[width=0.95\columnwidth]{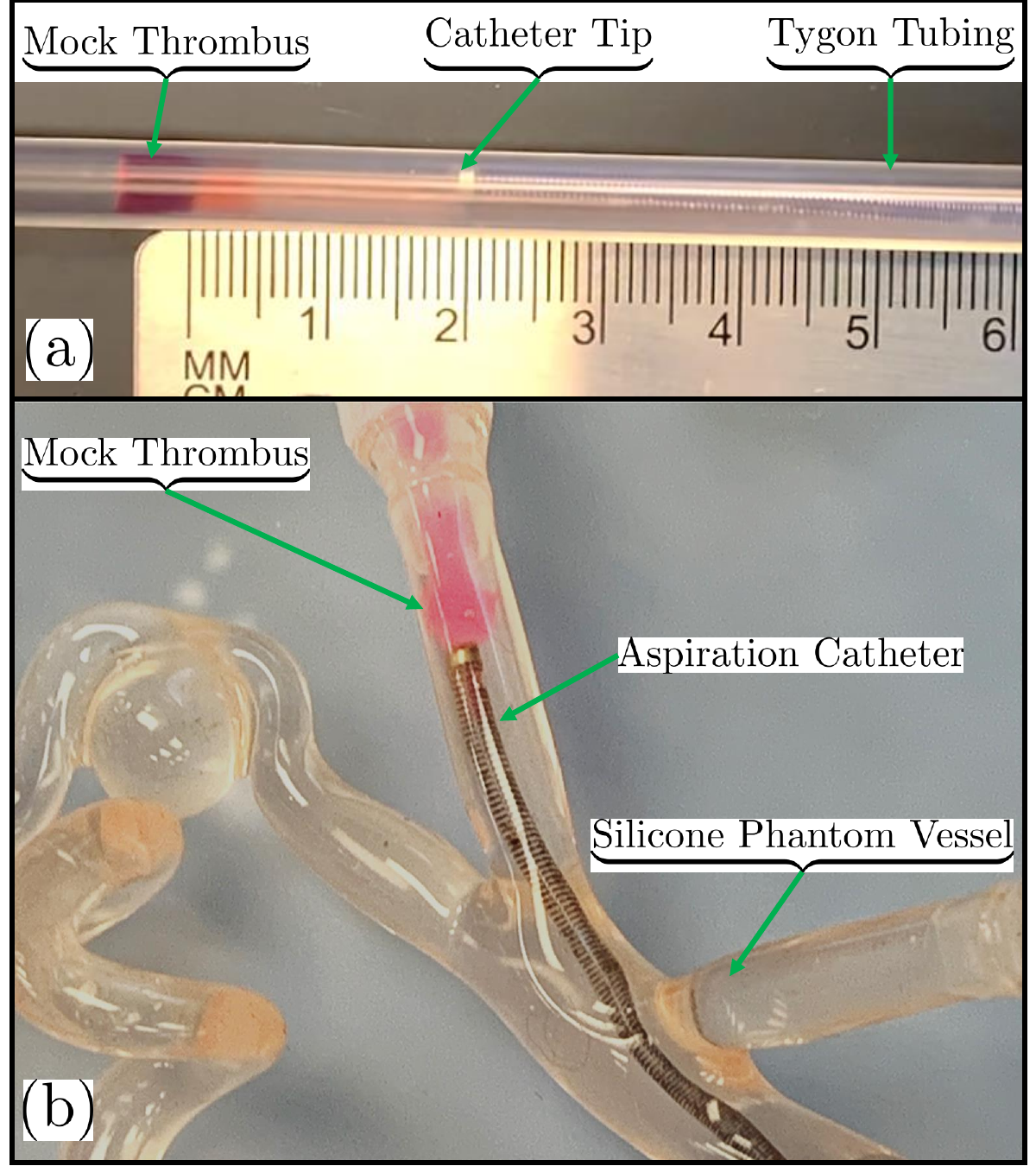}
        \caption{Phantom Models including vessels and mock thrombi. (a) shows an aspiration catheter inserted in a single Tygon tube vessel with a mock thrombus made from TrueClot synthetic clot material. (b) shows the aspiration catheter inserted in a branch of the United Biologics Neuro System Trainer, with a noise putty used for mock thrombus.}\label{fig:phantom_clots}
\end{figure}
\subsection{Phantom Models} \label{sec:models}
\par Clinically relevant vascular phantom models were used to collect data to train and validate the SVM classification model. Fig.~\ref{fig:phantom_clots}(a) shows a meter-long straight vessel phantom made of a thermoplastic elastomer (TPE) tube (Saint-Gobain, Tygon S3 B-44-4X) having an internal diameter of 3.2mm (1/8") and sealed with a catheter sheath introducer on the proximal end. The distal end of the tubing was blocked by a  mock thrombus (TrueClot, Blood Simulant and Clotting Solution) with a slightly oversized rod placed behind the mock thrombus to keep it fixed in place while maintaining a watertight seal. This phantom occluded vessel was filled with water to simulate blood so the catheter tip could be visualized within the model. The catheter was inserted into the sheath introducer while maintaining a seal, and the valve on the introducer was left open, such that, as the catheter was inserted, water could displace to maintain the overall pressure within the vessel. This phantom model was used for training tasks (Sec.~\ref{sec:algorithm}) since it represented the simplest model for a fluid-filled vessel with a soft thrombus at one end.
\par Fig.~\ref{fig:phantom_clots}(b) shows the second, more clinically relevant, phantom setup based on a  silicone arterial simulator including cerebral vessels (United Biologics, Neuro System Trainer), through which pulsatile blood flow mimicking the cardiac rhythm is introduced by a circulating pump (United Biologics, Flowtek 125). This model was used for robust verification of the clot detection model (Sec.~\ref{sec:experiments}) with anatomical relevant artifacts, including branching vessels, pulsatile flow, and tortuous vessel paths. It was also used for the user study (Sec.~\ref{sec:userstudy}), as it enables contrast flow for DSA images. The mock thrombus material used in this model was a noise putty (JA-RU, Flarp!), which has been previously demonstrated as a useful phantom thrombus model in \cite{nemoto2021pull}. The TrueClot mock thrombus was not used for this phantom model since it was more difficult to control the clot placement with TrueClot than the noise putty in the variable geometry of this model. Additionally, the noise putty mock thrombus validates that the model is robust to different clot material properties because it was trained with a different clot material.

\subsection{Detection Algorithm} \label{sec:algorithm}
\par Flow in an aspiration catheter can be described as Hagen-Poiseuille flow to describe the pressure loss over the length of the catheter. In the case of the unoccluded catheter, the change in pressure from the catheter tip to the catheter base (syringe) is a function of the flow resistance and the displaced fluid flowrate from the syringe stroke:
\begin{equation}\label{eq:HagenPoiseulle}
    \Delta p = \frac{128\eta L}{\pi D^{4}}Q
\end{equation}
where $\Delta p$ refers to pressure drop across the catheter. The fraction term refers to the flow resistance in a catheter system, where $\eta$ refers to the viscosity of blood, $L$ is the length of the catheter, and $D$ is the inner diameter of the catheter. The flow rate in the catheter, $Q$, is defined by the syringe oscillation of 0.4mL at 4Hz.
\par For multiple unoccluded samples, the pressure drop should be similar as the flow resistance of the system remains uniform. However, when an occlusion prohibits flow at the tip of the catheter, the flow resistance of the system is disturbed, which greatly increases the pressure drop along the catheter length. The detection algorithm will leverage this effect by classifying whether or not the pressure drop corresponds to thrombus contact or no thrombus contact. The first feature used in this classification algorithm, referred to as the relative average pressure, will serve to bias the sampled signal from a reference signal, which is collected away from the thrombus in open vessel. For a given collection sample, $\mathbf{p}_i$, taken away from a reference signal, $\mathbf{p}_{ref}$, this feature would be computed by:
{\color{black}
\begin{equation}{\label{eq:feat1}}
    \bar{\mathbf{p}}_{rel} = \bar{\mathbf{p}}_{i} - \bar{\mathbf{p}}_{ref}
\end{equation}}
where $\bar{\mathbf{p}}_{rel}$ refers to the average value over the sample signal. The second feature used for classification, referred to as the pressure change from prior, reflects a sudden change in pressure drop from a previous reading. This feature is computed by:
\begin{equation}{\label{eq:feat2}}
    \Delta \bar{p} = \bar{\mathbf{p}}_{i} - \bar{\mathbf{p}}_{i-1}
\end{equation}
where $\mathbf{p}_{i-1}$ refers to the preceding collection pressure signal.
\par The detection algorithm employs a support vector machine (SVM) classification model with a Gaussian kernel to distinguish catheter-clot contact from no contact. SVM was chosen above other machine learning models for its ability to maximize the distance from the classification hyperplane (therefore achieving the best margin of separation between groups \cite{Vapnik1998_support_vector_methods}) and its simplicity of implementation in real-time control. Since the data was not linearly separable, a Gaussian kernel was chosen to determine the boundary. A supervised learning approach was used to train the model. The SVM classification algorithm was implemented using Matlab's Statistics and Machine Learning Toolbox and was run on the host computer described in Section~\ref{sec:system}. 
\begin{figure}
    \centering
    \includegraphics[width=0.95\columnwidth]{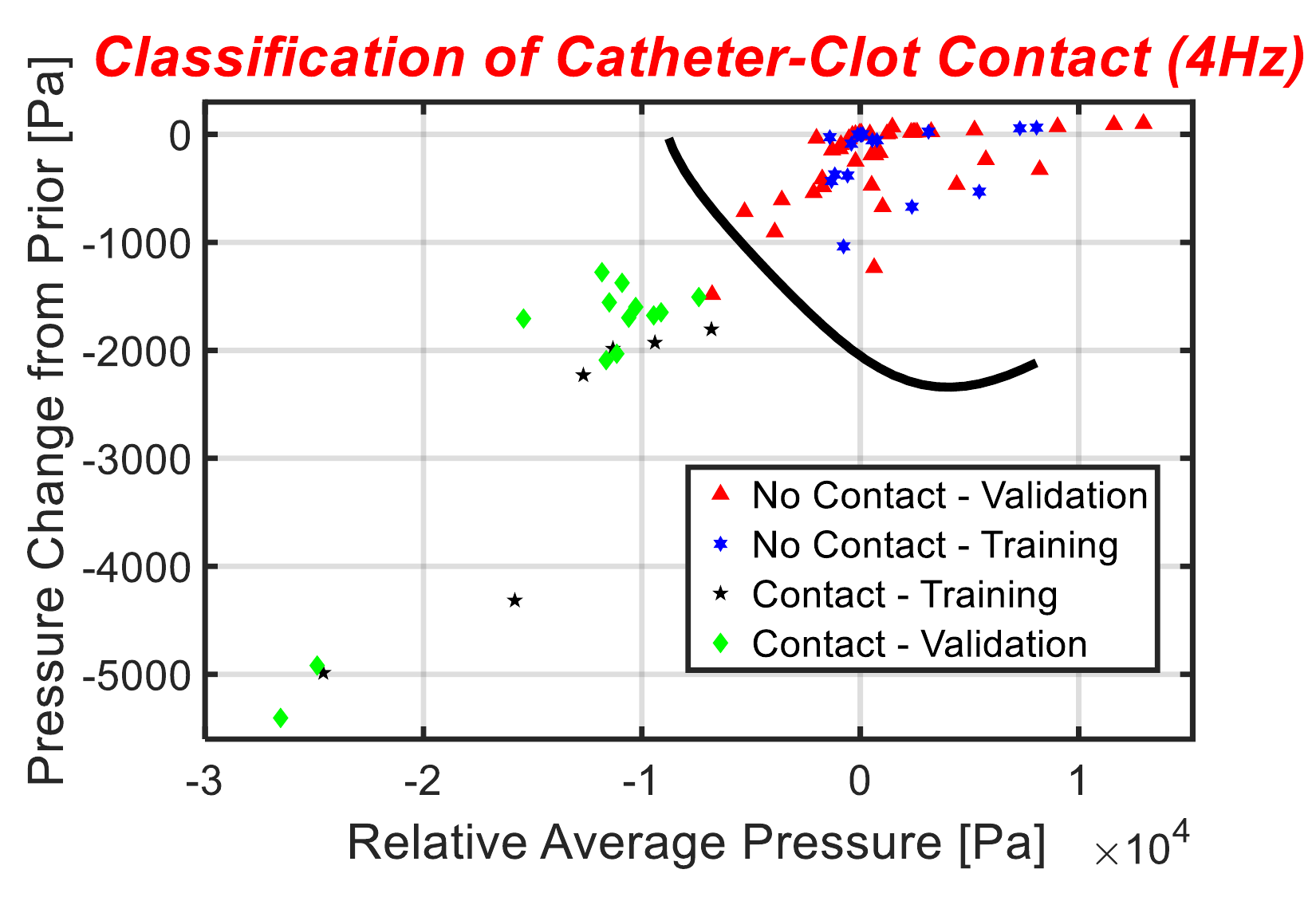}
    \caption{An example validation of the model using 30\% of the data to train and 70\% to validate.}\label{fig:training}
\end{figure}
\subsubsection{SVM Training and Validation}
\par Labeled data was collected using the first setup described in Section \ref{sec:models}. The catheter (Medtronic, React 71) was inserted into the clear mock vessel and pressure waves were measured at distances of 150mm, 20mm, 10mm, 5mm, and in contact (0mm) with respect to the thrombus. Data used to train the models were collected over 19 trials that each gave 4 data points, resulting in a total of 76 labeled points (57 labeled as the catheter was not in contact with the clot, and 19 labeled as the catheter was in contact with the clot). From the raw pressure wave, relative average pressure and pressure change from prior were used as input to the SVM. These two features are defined above and are shown in Alg.~\ref{alg:svm_algo} on lines 8-9, respectively.
\par For model validation, a random sample of 23 data points (30.3\%) were used to train the SVM model and 53 (69.7\%) were used to validate the models. The classification results of one such random sample are reported in Fig.~\ref{fig:training}. These random samples were computed ten separate times, resulting in average accuracy and F1 scores of 97.17\% and 94.66\%, respectively. While this sample of data indicated that the model extrapolates well to data outside of the training set, given the small sample size of collected data (n=76) it is desirable to use as much of this data as possible for training. Therefore, the model was regenerated with all 76 collected points, and was validated using 10-fold cross-validation with a holdout again of 70\% of data for testing. The resulting classification loss was 0.0189, indicating that the model was considerably accurate throughout the full data set. 
\subsubsection{Algorithm Implementation}
\par The model was then implemented into an interactive MATLAB R2016a (Mathworks) script that interfaces with the real-time control computer (Sec.~\ref{sec:system}) and predicts whether unlabeled data is in contact or not in contact with the clot (Algorithm~\ref{alg:svm_algo}). The program is started once the catheter has been inserted at an arbitrarily far distance from the clot, at which point the reference measurement is taken. The following process is then repeated until contact is detected. The algorithm first prompts the user to indicate when they are ready to conduct the next measurement. Once input is received, the syringe is oscillated at a stroke of 0.4mL at 4Hz for 2 seconds while collecting the raw pressure wave from the pressure sensor. The two input features are then extracted from the raw data and fed into the model to predict whether the catheter is or is not in contact with the clot. An audible tone and a visual message reports the result of the prediction to the user. 
\begin{algorithm}
\caption{Thrombus Detection Algorithm}
\label{alg:svm_algo}
\begin{algorithmic}[1]
\STATE{Collect reference pressure profile}
\STATE $\mathbf{p}_{ref} \gets \text{reference pressure profile}$
\STATE $\mathbf{p}_{prior} \gets \mathbf{p}_{ref}$
\STATE{Contact = False}
\WHILE{Contact is False} 
\STATE{Collect pressure profile}
\STATE $\mathbf{p}_{current} \gets \text{pressure profile}$
\STATE $\bar{p}_{rel} = \bar{\mathbf{p}}_{current} - \bar{\mathbf{p}}_{ref}$ 
\STATE $\Delta \bar{p} = \bar{\mathbf{p}}_{current} - \bar{\mathbf{p}}_{prior}$ 
\STATE{Contact = predictContact($\bar{p}_{rel}$,~$\Delta \bar{p}$)}
\IF{Contact is True}
\STATE {Provide auditory feedback signal}
\PRINT Contact achieved!
\STATE {\textbf{break}}
\ELSE 
\STATE{Auditory Feedback of Contact Not Achieved} 
\STATE $\mathbf{p}_{prior} \gets \mathbf{p}_{current}$
\ENDIF
\ENDWHILE
\end{algorithmic}
\end{algorithm}
\subsection{Benchtop Validation of Detection Robustness to Vessel Geometry and Heartbeat}\label{sec:expmethod}
Section \ref{sec:algorithm} presented the initial feasibility of contact detection using SVM with training data obtained using a straight mock blood vessel as in Fig.~\ref{fig:phantom_clots}(a). To test whether the SVM classification using the training data generalizes to successfully detect clot contact for realistic vascular geometry and in the presence of heartbeat flow excitation, we used the United Biologics phantom model shown in Fig.~\ref{fig:sensing_system}(a) and Fig.~\ref{fig:phantom_clots}(b).   
\par Experimental validation in this phantom model consisted of placing mock thrombi in ten different vessel locations across the model. These ten different vessel locations were selected to force the aspiration catheter to conform to highly variable shapes, ensuring that detection performance is robust to catheter shape uncertainty. While LVO's predominantly are located in the ICA and MCA, additional and less clinically relevant clot locations were used as extreme use cases of the approach. At each of these ten locations, fifteen trials of clot detection were performed, with each trial attempting four detection samples: three samples with the catheter not in contact, and one with the catheter in contact. In total, there were over 600 detection samples, since some trials included extra samples. One third of the data was collected with no heart rate and constant flow, another third was collected with a heart rate of 70 beats per minute (bpm), and the remaining third was collected with heart rate of 100bpm. The results of this evaluation are presented in Section \ref{sec:experiments}.
\subsection{User Study Protocol}\label{sec:studymethod}
\par A user study was designed to test the impact of our sensing technology on the likelihood that an experienced neuro-interventionalist would correctly detect thrombus contact with the catheter tip. The user study protocol was approved by the Vanderbilt University Institutional Review Board (IRB) \#210694 on January 17th, 2023. The user study consisted of users with neuroendovascular training (four fellows and one attending neurointerventionalist). ~Statistical significance and study design considerations are presented in Section~\ref{sec:userstudy}. This section focuses on presenting the detailed setup and experimental protocol followed by each study subject. 
\par The experimental setup for this study includes the same vascular phantom model from United Biologics (used for benchtop validation in the preceding section, Sec. \ref{sec:expmethod}). This setup was placed on a surgical bed under bi-plane fluoroscopic imaging using Siemens Artis-Q and  Philips Allura XPER FD20/20 fluoroscopy machines in two separate neurological operating suites which were randomly assigned to users based on availability of the operating suites and the users (study subjects). A clot, prepared in an identical manner as described in Section~\ref{sec:expmethod}, was introduced into the MCA during each experiment. The catheter tip distance to the clot was also observed using a camera placed to directly visualize the clot and catheter tip.  
\par Each study subject repeated a trial consisting of navigation and clot detection. For each trial, a mock thrombus was placed in the MCA of the phantom model, and a separate expert neurointerventionalist obtained the digital subtraction angiogram (DSA). Once the DSA was collected, the user was invited into the room to view the angiographic roadmap and they were asked to use their judgment to estimate the thrombus location on the roadmap image. The user was then asked to detect the clot under two study conditions: 1) \textbf{``control condition"}, in which the user only leveraged fluoroscopic imaging, and 2) \textbf{``sensing condition"}, in which the user was provided auditory feedback from the detection algorithm in addition to fluoroscopic imaging. In both study conditions the users were asked to advance the catheter while starting with an estimated distance of 10mm away from the clot based on their judgment when observing the fluoroscopy images. They then advance the catheter to an estimated location 5mm away from the clot and another location 0mm away from the clot. For each one of these three estimated distances (10mm, 5mm and 0mm) they pause and verbally declare whether they think the clot is in contact or not and, for the sensing study condition, our sensory modality was also used to produce a concurrent declaration based on the output of the SVM classification. These matched user and algorithm contact declarations are collated and used in Section~\ref{sec:userstudy} to produce data for the confusion matrix \cite{fawcett2006confusion_matrix} associated with quantifying the classification performance.    
\par Since users were asked to declare their contact estimate during the sensing condition, this provided an additional data set that allows us to judge whether the use of the sensory excitation affected the user's prediction of contact or introduced an experimental bias. In the following text, we refer to the data set collected during these experiments as the \textbf{``declarative condition"}. 
\par Each user was asked to repeat 12 trials with the clot locations changed between each trial to prevent users from remembering the clot location. In addition, the order of trial conditions was randomized to ensure that six control (three left MCA, three right MCA) and six sensing (three left MCA, three right MCA) trials were conducted with minimal systematic study bias. After each study, the results were collated with ground truth images to determine classification results for the control and sensing trials of each user. 
\par For both study conditions, users were asked to place the catheter tip approximately 10mm, 5mm, and 0mm away from the surface of the clot. This provided two non-contact, and one contact data points per trial (although some cases provided more/less data points depending on the user's judgment approximating the thrombus). In the control condition, once the user reached each of the estimated distances, they would pause and indicate which position they were in, and a ground truth camera image was recorded to determine whether or not the catheter was in contact with the thrombus. In the sensing condition, the same pause and ground truth camera image was recorded, but with each pause at the estimated distance the detection algorithm was conducted and the result was reported back to the user as auditory feedback. The results stemming from this user study are included in Section \ref{sec:userstudy}.
\section{Results} \label{sec:results}
\begin{figure}[htbp]
        \centering
        \includegraphics[width=0.9\columnwidth]{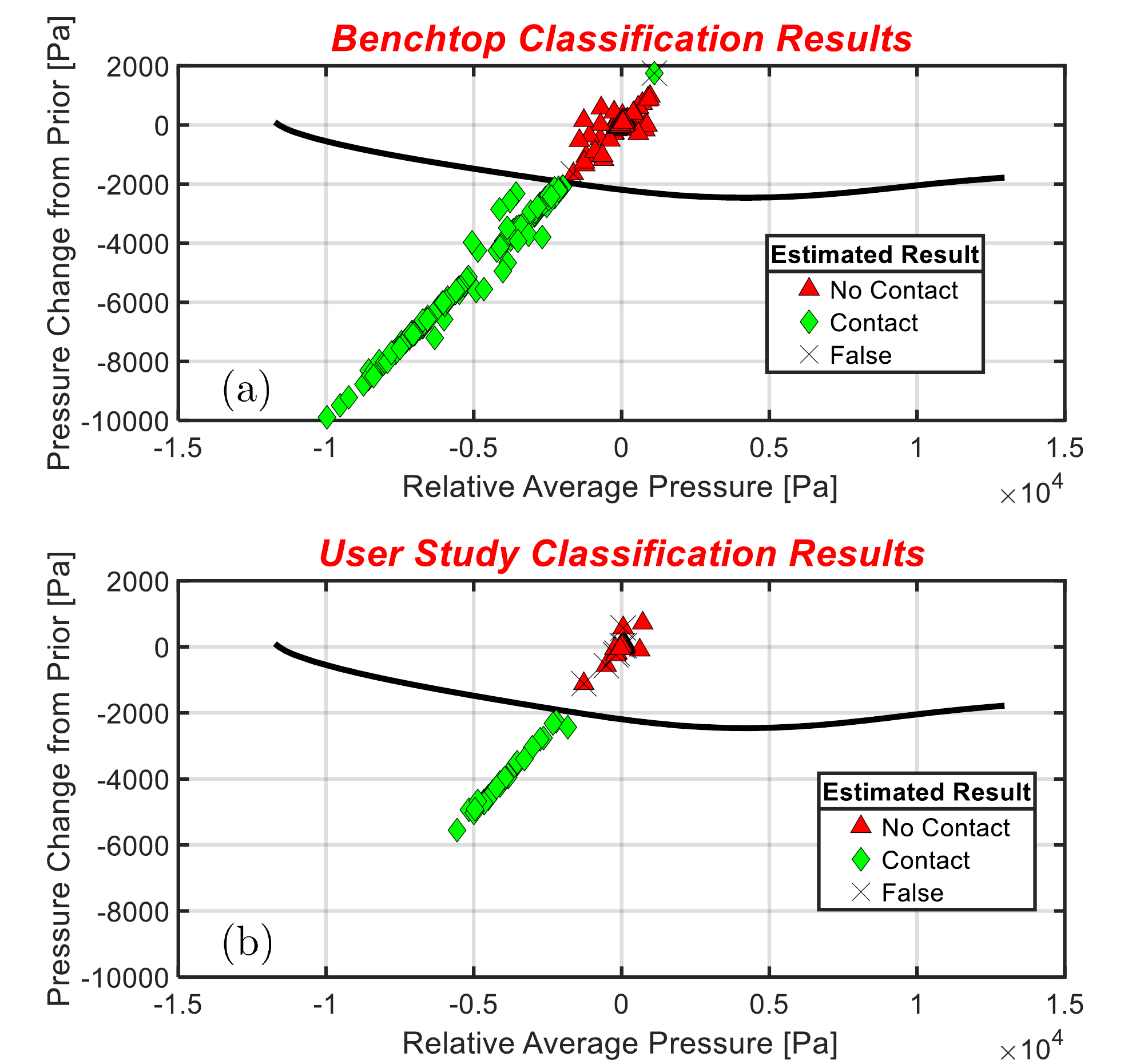}
        \caption{Classification results on clinically relevant phantom model for (a) benchtop validation and (b) the user study. Both plots show feature-space plotting of classification results around the support vector boundary. In both plots, the diamonds represent instances of estimated contact, while the triangles represent estimated non-contact. Incorrect estimates are marked with cross symbols.}\label{fig:classification_experiments_user_study}
\end{figure}
\subsection{Benchtop Validation Results} \label{sec:experiments}
\begin{figure*}[htbp]
        \centering
        \includegraphics[width=0.88\textwidth]{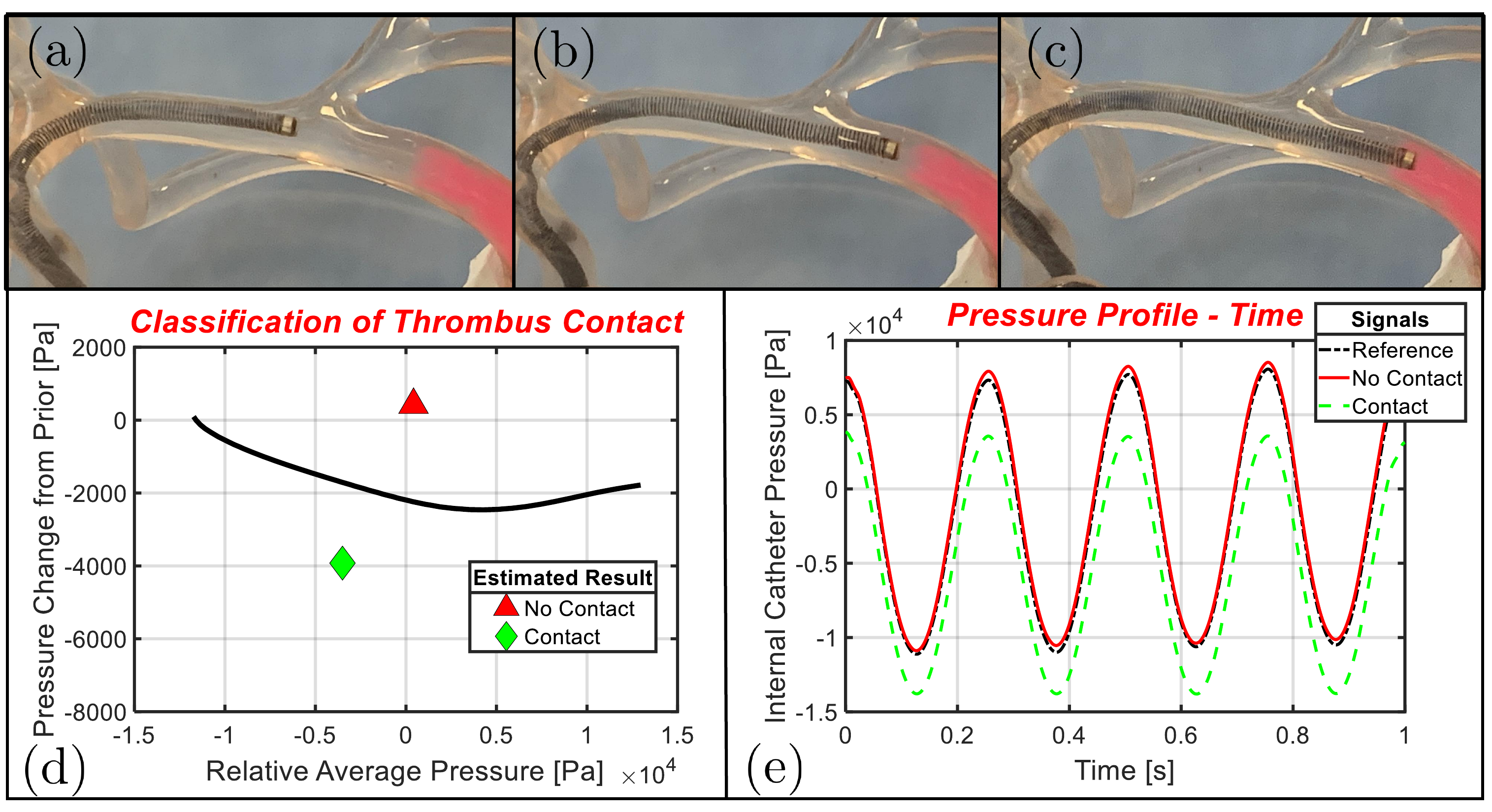}
        \caption{From one sample collection: (a-c) images of a reference, non-contact, and contact sample. (d) presents feature-space classification results for this trial, and (e) shows the pressure profile over collection time.}\label{fig:sample_trial}
\end{figure*}
\par In total, 606 detection samples were collected across the ten clot locations. Among this total, 456 samples were collected with the catheter not in contact with the clot, and 150 samples were collected with the catheter in contact with the clot. In only two samples (out of 606 samples) were false detection results recorded (one false negative, one false positive). Thus, the percentage error of this validation was $0.33$\%, proving that the proposed algorithm achieved an accuracy of $99.67$\%. Additional classification measures are included in Table \ref{tab1}. Each data point represented in feature-space as shown in Fig. \ref{fig:classification_experiments_user_study}(a), where the contour line represents the boundary separating the Contact and Non-Contact regions as defined by the classification model training in Section \ref{sec:algorithm}. Results for one particular trial are included in Fig.~\ref{fig:sample_trial}, where (a-c) show the catheter in position for a reference, no contact, and contact signal, respectively. At each of those positions, the classification results and pressure profiles are shown in Figs.~\ref{fig:sample_trial}(d-e).
\begin{table}
\centering
\caption{Classification Results for Benchtop Validation on Silicone Phantom Model}
\label{table}
\setlength{\tabcolsep}{5pt}
\begin{tabular}{|p{100pt}|p{50pt}|}
\hlineB{4}
\textbf{Measure}& \textbf{Value} \\
\hlineB{4}
Accuracy & 
0.9967 \\
\hline
Precision & 
0.9933 \\
\hline
Recall (Sensitivity) &
0.9933 \\
\hline
Specificity &
0.9978 \\
\hline
F1 Score &
0.9933 \\
\hline
F2 Score &
0.9933 \\
\hlineB{4}
\end{tabular}
\label{tab1}
\end{table}
\subsection{User Study Results}\label{sec:userstudy}
\begin{figure*}[htbp]
        \centering
        \includegraphics[width=0.99\textwidth]{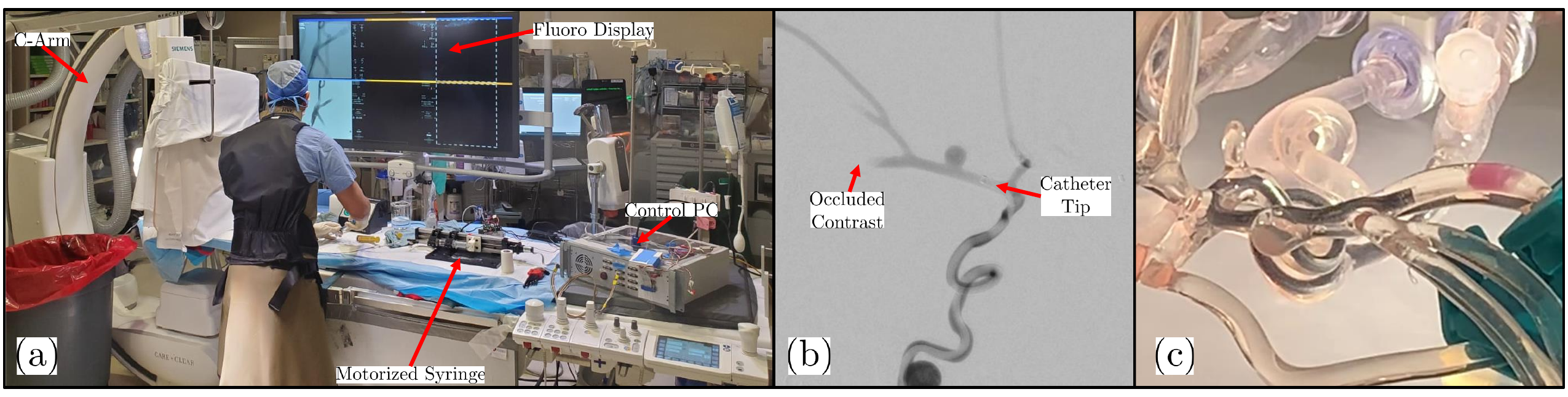}
        \caption{Components of the user study: (a) the setup from Fig. \ref{fig:sensing_system}(a) placed on the OR bed with interventionalist, (b) a DSA indicating a thrombus in the M2 segment of the MCA, (c) a ground truth image of the same M2 segment with the catheter tip approaching the mock thrombus. }\label{fig:OR_setup}
\end{figure*}
\begin{table*}[!b]
\centering
\caption{User Study Classification Results. Numbers between brackets indicate 95\% confidence intervals. Odds ratios are reported relative to the control condition.}\label{tab:userstudydata}
\setlength{\tabcolsep}{0pt}
\begin{tabularx}{0.99\textwidth}{ 
  | >{\centering\arraybackslash}X 
  | >{\centering\arraybackslash}X 
  | >{\centering\arraybackslash}X
  | >{\centering\arraybackslash}X 
  | >{\centering\arraybackslash}X 
  | >{\centering\arraybackslash}X 
  | >{\centering\arraybackslash}X 
  | >{\centering\arraybackslash}X 
  | >{\centering\arraybackslash}X 
  | >{\centering\arraybackslash}X 
  | >{\centering\arraybackslash}X 
  | >{\centering\arraybackslash}X 
  | >{\centering\arraybackslash}X 
  | >{\centering\arraybackslash}X 
  | >{\centering\arraybackslash}X |}
 \hlineB{4}
 \multicolumn{4}{|c|}{\textbf{Control Condition}} & \multicolumn{4}{|c|}{\textbf{Declarative Condition}} & \multicolumn{4}{|c|}{\textbf{Sensing Condition}} \\
 \hlineB{4}
 \multicolumn{2}{|c|}{} & \multicolumn{2}{c|}{Estimated} & \multicolumn{2}{c|}{} & \multicolumn{2}{c|}{Estimated} & \multicolumn{2}{c|}{} & \multicolumn{2}{c|}{Estimated} \\ \cline{3-4} \cline{7-8} \cline{11-12}
 \multicolumn{2}{|c|}{} & Contact & No Contact & \multicolumn{2}{c|}{} & Contact & No Contact & \multicolumn{2}{c|}{} & Contact & No Contact \\
 \hline
 \multirow{ 2}{*}{Actual} & Contact & 19 (21.1\%) & 4 (4.4\%) & \multirow{ 2}{*}{Actual} & Contact & 14 (16.3\%) & 6 (7.0\%) & \multirow{ 2}{*}{Actual} & Contact & 29 (27.1\%) & 7 (6.5\%) \\ \cline{2-4} \cline{6-8} \cline{10-12}
 & No Contact & 11 (12.2\%) & 56 (62.2\%) & & No Contact & 12 (13.9\%) & 54 (62.8\%) & & No Contact & 0 (0\%) & 71 (66.4\%) \\
\hline
\multicolumn{2}{|c|}{Combined error rate} & \multicolumn{2}{c|}{15/90 (16.7\%)} & \multicolumn{2}{c|}{Combined error rate} & \multicolumn{2}{c|}{18/86 (20.9\%)} & \multicolumn{2}{c|}{Combined error rate} & \multicolumn{2}{c|}{7/107 (6.5\%)} \\
\hline 
\multicolumn{2}{|c}{ } & \multicolumn{2}{c|}{} & \multicolumn{2}{c|}{Odds ratio} & \multicolumn{2}{c|}{\parbox{2.8cm}{\centering \vspace{2pt} 0.75 [0.34, 1.62] \newline $p$=0.466 \vspace{2pt}}} & \multicolumn{2}{c|}{Odds ratio} & \multicolumn{2}{c|}{\parbox{2.8cm}{\centering \vspace{2pt} \textbf{2.86 [1.09, 7.44] \newline \textbf{$\boldsymbol{p}$=0.031}} \vspace{2pt}}} \\
 \hlineB{4}
\end{tabularx}
\end{table*}
\par Figure~\ref{fig:OR_setup} shows the setup of the user study, with Fig.~\ref{fig:OR_setup}(a) showing the interventionalist interacting with the setup and fluoroscopic imaging. Figs.~\ref{fig:OR_setup}(b-c) show the fluoroscopic image and ground truth camera image for a particular sample collection, respectively. In total, five users each conducted 12 trials of the protocol resulting in a total of 283 data points. These 283 data points consisted of 90 data points from the control condition (fluoroscopy alone), 107 data points from the sensing condition (fluoroscopy + auditory feedback from our classifier), and 86 data points from the declarative condition. These 86 data points were included to ensure that any differences observed between the control and sensing conditions were not an accidental byproduct of the experimental setup.
\par Table \ref{tab:userstudydata} includes all data points from the user study. When performing under the control condition, the users incorrectly reported catheter contact with the mock thrombus 16.7\% of the time (n=15, 4 False Negative, 11 False Positive). The data set for the declarative condition included 86 user estimates where 20.9\% were incorrect  (n=18, 6 False Negative, 12 False Positive). In contrast to these results, when users performed under the sensing condition, only 7 (6.5\%) incorrect instances were observed, and all classification results are reported in Fig. \ref{fig:classification_experiments_user_study}(b). Each of those incorrect values was when contact was achieved, but the system did not detect contact (False Negative).
\par Given the lack of total independence in the data values (i.e., 5 users conducted multiple trials to obtain the 283 data points), a mixed-effects logistic regression model was used to test the null hypothesis of no statistically significant difference in the contact detection accuracy between the study conditions. In that analysis, not only were the standard errors adjusted for the lack of independence in the data, 'user' was specified as a random effect to test whether differences among users may be generating any primary condition effects. 
\par There was a statistically significant effect of study condition (Wald $\chi^{2}$=7.88, p=0.0195). Specifically, that effect was due to the improvement in accuracy in the sensing condition compared to the control condition (odds ratio\footnote{The odds ratio (OR) is the chance in correctly detecting contact divided by the chance of the same when performing under the control condition.} OR=2.86, 95\% C.I. = $\left[1.09, 7.44\right]$, p=0.031). There was no statistically significant difference between the declarative condition and control condition (OR=0.75, 95\% C.I.=$\left[0.34, 1.62\right]$, p=0.466), indicating that the experimental setup was consistent between study conditions. Furthermore, there was no demonstrated effect of the user on accuracy (LR $\chi^{2}<$0.01, p$>$0.999).
\section{Discussion} \label{sec:discussion}
\subsection*{Benchtop Experimental Validation}
\par The purpose of the benchtop experimental validation was to determine not only the accuracy of the proposed sensing modality, but also to examine the sensitivity of the sensing capability with respect to clinically relevant variables, including vessel tortuosity, vessel size, and heart rate.
\par With a successful thrombus detection rate of 99.67\% across the experimental validation, the vacuum sensing modality was found to be effective in 10 different highly variable tortuous paths. While constant flow rate (i.e. no heart rate), 70 bpm and 100 bpm heart rates were examined, there seemed to be no effect on the efficacy of the detection method. 
\par This sensing technique requires the clot to be contacting the annular tip of the catheter, such that as vacuum is applied the clot would be drawn into the catheter lumen. If, however, the clot is making contact only with the outside of the catheter body and not directly with the catheter tip, it would be expected that the sensing algorithm would not detect contact. This was correctly observed in multiple benchtop trials. For each contact trial in the benchtop experimental validation, it could be observed that the thrombus was drawn into the catheter tip with the pulsatile vacuum excitations, indicating that contact corresponds to the ability to aspirate the thrombus. 
\subsection*{User Study}
\par This user study was designed to test this proposed sensing method in a realistic clinical workflow and to compare the experts' ability to correctly detect contact with and without this sensory feedback.
\par The results of this user study clearly indicate that neurointerventionalists are significantly more likely to accurately detect contact with a thrombus with the proposed sensing modality than they are relying purely on fluoroscopic imaging alone. In order to ensure that this result was purely a function of the sensing modality, and was not biased by the experimental workflow, the user estimates during the declarative condition were compared to the control condition and were found to be statistically similar. In contrast, results from the sensory condition were statistically different from the control condition and showed a drastic increase in the likelihood of correct clot detection (refer to the bold odds ratio cell in Table~\ref{tab:userstudydata}). 
\par After conducting the study, each user completed both a NASA TLX Survey to report on the mental workload of the study conditions, as well as an additional questionnaire. Every user indicated that having the additional sensing feedback in the form of an auditory cue was helpful to confirm or correct their visual assessment. Three of the five users indicated the mental burden of the task was reduced by including sensory feedback.
\par The proposed sensing modality is shown to enhance the interventionalists' ability to approximate catheter contact with a thrombus, while reducing the cognitive burden of relying on purely angiographic information, and represents an exciting opportunity to potentially improve FPE of mechanical thrombectomy by aspiration.
\subsection*{Limitations}
\par While most reasonable efforts were taken to make the detection method and experimental study as clinically relevant as possible, there were some limitation regarding materials used for data collection. The mock thrombus materials were used as they represented previously reported thrombus analogs, however their specific material properties may differ from coagulated human blood. Therefore, further training with varying clot material properties would improve the robustness of this approach. Additionally, our most clinically relevant phantom vessels were made of a silicone material which resembles but may not identically match typical artery material properties. While there may be differences between the material properties of the tested phantom models and diseased vessel, it is critical to note that the vessel wall stiffness does not significantly affect the vacuum profile when the catheter tip is in contact with the clot, since the clot-ingested catheter forms a closed system where the stiffness is dominated by the catheter walls. Furthermore, since the data samples are biased from a reference signal in the vessel, the effects of vessel stiffness are constant throughout sensing until contact is achieved.
\par When sensing contact at the distal tip of the catheter, the catheter may be either contacting the surrounding vessel, or the thrombus. It is critical that contact with the surrounding vessel is not misreported as contact with the thrombus. During benchtop verification experiments as well as the user study, multiple instances of catheter-wall contact could be observed, but which correctly detected no contact with a thrombus. For the system to misclassify catheter-wall contact as catheter-thrombus contact, the catheter tip would need to fully ingest a portion of vessel wall, which would require a sustained vacuum pressure significantly higher than the arterial pressure inside the vessel. However, as the recommended aspiration catheter diameter is around two thirds the occluded vessel diameter, it is unlikely the catheter would be able to turn sharply to make fully-occluding contact \cite{pampana2021tailored}. To more robustly prevent this false positive, the system can be controlled to limit the vacuum pressure applied at the oscillating syringe to be below a minimum diastolic pressure. By limiting this induced vacuum pressure at the catheter tip, the arterial pressure in the vessel would prevent the vessel wall from collapsing under the oscillating vacuum.
\par One of the aspects of the current workflow for the sensing modality includes taking a reference reading, and then stopping at discrete locations to perform the sensing algorithm for one second, before continuing deeper into the vessel. Users recommended the detection algorithm be trained to read a continuous stream of pressure data rather than taking discrete readings. This would more accurately coincide with the current thrombectomy workflow, and would further improve the efficiency of the procedure.
\subsection*{Future Works}
\par The next stage of validating this approach for use in humans must include testing feasibility in an animal (i.e. swine) model. This will begin to indicate whether or not anatomical differences between the phantom models (vessel and thrombus) used herein are similar enough to real anatomical tissue. 
\par While indicating classification of catheter contact with a thrombus is shown to be useful for the interventionalists, predicting catheter proximity to the thrombus would further enhance the interventionalists ability to understand the endovascular environment. This would require expanding from a classification model towards a regression model that will output predicted distance to the clot.
\section{Conclusion} \label{sec:conclusion}
\par In this work, we present a sensory modality which utilizes off-the-shelf aspiration catheters for detecting contact between the catheter tip and a thrombus. This method requires no additional sensors at the catheter's distal tip, rather utilizing a proximal extra-corporeal sensor to monitor internal pressure. By applying pulsatile vacuum excitation with a motorized syringe, catheter-thrombus contact can be predicted based upon the resulting pressure profiles. Benchtop validation of this approach yielded 99.67\% accuracy in detecting contact at different clot locations and heart rates. A user study demonstrated that providing contact sensory feedback to trained neurointerventionalists allowed them to more accurately locate thrombus than using fluoroscopic guidance alone. 

\bibliographystyle{IEEEtran}
\bibliography{main}
\end{document}